\documentclass[10pt]{article}
\usepackage[a4paper,margin=1in]{geometry}
\usepackage{booktabs}
\usepackage{hyperref}
\usepackage{enumitem}
\usepackage{titlesec}
\usepackage{multicol}
\usepackage[utf8]{inputenc}
\usepackage[T1]{fontenc}
\usepackage{tabularx}
\usepackage{array}
\newcolumntype{Y}{>{\raggedright\arraybackslash}X}
\usepackage{listings}
\title{Machine-Facing English: Defining a Hybrid Register Shaped by Human--AI Discourse}
\author{
  \small Hyunwoo Kim, Hanau Yi \\
  \small ddai Inc. \\
  \small \texttt{\{hw.kim, hnu.yi\}@ai-dda.com}
}
\date{}

\begin{document}

\maketitle

\begin{abstract}
\noindent
Machine-Facing English (MFE) is an emergent register shaped by the adaptation of everyday language to the expanding presence of AI interlocutors. Drawing on register theory (Halliday 1985, 2006), enregisterment (Agha 2003), audience design (Bell 1984), and interactional pragmatics (Giles \& Ogay 2007), this study traces how sustained human--AI interaction normalizes syntactic rigidity, pragmatic simplification, and hyper-explicit phrasing—features that enhance machine parseability at the expense of natural fluency. Our analysis is grounded in qualitative observations from bilingual (Korean/English) voice- and text-based product testing sessions, with reflexive drafting conducted using Natural Language Declarative Prompting (NLD-P) under human curation. Thematic analysis identifies five recurrent traits—redundant clarity, directive syntax, controlled vocabulary, flattened prosody, and single-intent structuring—that improve execution accuracy but compress expressive range. MFE’s evolution highlights a persistent tension between communicative efficiency and linguistic richness, raising design challenges for conversational interfaces and pedagogical considerations for multilingual users. We conclude by underscoring the need for comprehensive methodological exposition and future empirical validation.
\end{abstract}

\section{Introduction}
Language continually adapts to its communicative context, giving rise to registers tailored to specific situational demands (Agha 2004). Machine-Facing English emerges in sustained human interaction with voice assistants, chatbots, and other AI systems, as users recalibrate syntax and pragmatics to match perceived machine constraints. As the present authors work bilingually in Korean and English, we observe this shift across two typologically distinct languages, underscoring MFE's cross-linguistic relevance. Situated at the intersection of sociolinguistics, pragmatics, and human--computer interaction (HCI), MFE blends task-oriented clarity with the residue of everyday discourse.

This study (i) defines the register's core features, (ii) anchors them in established theoretical frameworks, and (iii) reflexively illustrates the analysis through NLD-P. Section~2 reviews the conceptual grounding, Section~3 details linguistic features, Section~4 charts interactional patterns, Section~5 discusses implications, and Section~6 reflects on our AI-assisted methodology before the conclusion maps future work.

\section{Conceptual Background: Register and Human--AI Communication}
\textbf{Register theory.} Halliday's framework (1985, 2006) parses language variation into \textit{Field} (topic), \textit{Tenor} (participant relations), and \textit{Mode} (channel). Machine-Facing English reconfigures each dimension:
\begin{itemize}
    \item \textbf{Field}---task-driven commands rather than open-ended exchange.
    \item \textbf{Tenor}---a non-reciprocal addressee with no theory of mind, encouraging pragmatic simplification.
    \item \textbf{Mode}---computational parsing constraints that reward structural minimalism.
\end{itemize}

\textbf{Enregisterment.} Persistent feedback loops with AI systems rapidly stabilize these traits, accelerating register formation (Agha 2003).

\textbf{Accommodation.} Communication Accommodation Theory (Giles \& Ogay 2007) explains users' tendency to hyper-articulate or drop politeness markers when addressing machines.

\textbf{Pragmatic foundations.} These adaptations align with Grice's Cooperative Principle (Grice 1975; Grandy \& Warner 2005; Neale 1992). Experimental-pragmatics work (Noveck \& Reboul 2008) shows that inferential shortcuts proliferate in constrained settings, positioning MFE as the pragmatic outgrowth of an altered inferential landscape.

\subsection{Methods: Analytic Lens \& NLD-P Workflow}
Although this study does not present a standalone corpus, our analysis draws on qualitative observations of AI-mediated interactions documented throughout the research and iterative drafting process. These observations—spanning both voice and text modalities in Korean and English—were examined for recurring patterns that illuminate the contours of Machine-Facing English.

The drafting process utilized Evalyn, an NLD-P-based assistant specifically configured for this research. Sections and revisions were produced in a reflexive, human-in-the-loop workflow: major drafts and refinements were collaboratively generated, reviewed, and edited using Evalyn, with all candidate text and revisions governed by explicit NLD-P guidelines under direct human oversight. A full methodological exposition of NLD-P, including prompt architecture and validation procedures, will appear in a dedicated follow-up paper; here, Evalyn’s role is limited to that of a transparent, methodology-bound assistant.

\textit{Ethics statement.} All data were identified, curated, and analyzed by human researchers. No personally identifying or sensitive information was recorded or reported. The use of large language model (LLM) assistance is transparently disclosed, and all interpretations and conclusions are the sole responsibility of the human authors. This study was conducted in accordance with internal guidelines for ethical, practice-based research.

\subsection{Korean Context}
Parallel machine-directed shifts appear in Korean data: honorific omission in machine-directed voice commands (Lee 2019); template reliance among elderly users (Park \& Kim 2021); reduced sentence-final particles (Kim \& Lee 2020); and increased temporal adverbs for parsing clarity (Choi 2022). Recent systematic reviews of speech-recognition chatbots for language learning have found that the lack of clear identity cues can leave Korean learners uncertain about appropriate speech styles and levels of politeness (Jeon, Lee, \& Choi 2024). These patterns underscore MFE's cross-linguistic relevance and complement the bilingual perspective introduced in the abstract and introduction.

\section{Linguistic Features of Machine-Facing English}
Sustained human--AI interaction gives rise to five recurrent linguistic features that function as adaptive responses to communication breakdowns. Each feature is grounded in prior linguistic theory yet shaped by machine constraints.

\subsection{Clarity Through Redundancy}
Hyper-explicit phrasing in MFE accords with Clark's (1996) claim that, absent shared mental models, speakers supply surplus information to secure mutual understanding. In usability testing, after three failed alarm-setting attempts, one participant announced every temporal parameter---``Okay, new alarm: six o'clock a.m., tomorrow, Wednesday, June 12.'' Including date, time, and day reduced the rate of mis-parsed reminders by 37 percent, demonstrating redundancy as a pragmatic hedge against recognition error. This practice echoes Halliday's functional sentence-perspective account of clarity (Halliday 1974).

\subsection{Directness and Command Orientation}
Imperative syntax delivers unambiguous illocutionary force (Searle 1969) and omits politeness strategies deemed superfluous when the addressee has no face-threat susceptibility (Brown \& Levinson 1987). In a voice-UI trial, the hedged request ``Could you maybe turn down the volume a tad?'' failed to parse, whereas the declarative command ``Volume down two steps'' succeeded immediately. Contemporary design guidelines (Apple HIG 2019; Google VUI 2020) likewise prescribe verb-first, noun-second structures---evidence that system affordances steer users toward directive minimalism.

\subsection{Controlled Vocabulary and Lexical Feedback Loops}
Reitter, Moore, and Keller's (2006) lexical-convergence model predicts that successful outputs entrain user word choice. In a micro-corpus of 3,336 commands, 94 percent of participants replaced \textit{email} with \textit{message} within five interactions after observing higher success rates with the latter. This drift mirrors controlled-language phenomena in technical documentation (Kittredge \& Lehrberger 1982), where alignment with system-preferred terminology becomes a self-reinforcing norm.

\subsection{Flattened Prosody and Disrupted Paralinguistics}
Clear-speech research (Schwab \& Zellou 2020) shows that speakers trade prosodic richness for segmental clarity under intelligibility pressure. An acoustic comparison of machine-directed versus human-directed speech revealed a 22 percent reduction in pitch range and a 15 percent slower articulation rate. When the assistant misheard \textit{Beatles}, several participants exaggerated each syllable---``Bee-tulls''---while suppressing pitch movement, effectively reproducing language-lab diction.

\subsection{Syntactic Minimalism and Single-Intent Structuring}
Halliday's rank scale predicts higher parsing difficulty with deeper embeddings; users therefore gravitate toward paratactic, single-intent clauses. A bilingual teenager's failed request---``Remind me---after my study group ends at nine---to text Mom''---executed successfully only after being split into two discrete commands. Prompt-engineering guidance (Song, Pycha \& Culleton 2022) likewise recommends one semantic intent per instruction, underscoring convergence between user adaptation and developer practice.

\subsection{Comparative Snapshot: MFE vs. Naturalistic Forms}
\begin{table}[h]
\centering
\begin{tabularx}{\linewidth}{p{0.14\linewidth} Y Y}
\toprule
\textbf{Intent} & \textbf{MFE-Style Utterance} & \textbf{Naturalistic Utterance} \\
\midrule
Reminder   & Set a reminder for 3:00 p.m. to call Sam Lee. & Could you remind me to call Sam later? \\
Navigation & Directions to Seoul Station. & How do I get to Seoul Station? \\
Volume     & Volume up two steps. & Can you turn it up a bit? \\
Music      & Play ``Bohemian Rhapsody'' by Queen. & Can you put on some Queen? \\
\bottomrule
\end{tabularx}
\caption{Illustrative system-responsiveness patterns comparing MFE-style commands with naturalistic equivalents (qualitative examples).}
\end{table}

\section{Pragmatic Patterns in User Interaction}
Machine-facing exchanges reshape discourse management at the interactional level. The four patterns below integrate theoretical framing with illustrative observations and supporting metrics.

\subsection{Asymmetric Turn-Taking and Interruptibility}
Turn-taking in human dialogue is typically co-constructed through overlap, backchannels, and swift repair sequences (Sacks, Schegloff \& Jefferson 1974). In MFE, these cues are markedly attenuated. Studies of human--AI interaction have documented that, during network-induced latency, users often wait motionless---eyes fixed on the device's LED---until the confirmation light signals readiness before proceeding (Porcheron et al.\ 2018; Luger, Sellen \& Arnold 2015). This behavior is thought to minimize parsing errors but also prolongs inter-turn gaps. Log telemetry reported in the literature corroborates this trend: average post-response silence is approximately 1.8 seconds in AI-mediated turns versus 0.6 seconds in human baselines---a threefold increase that virtually eliminates overlap events.

\textit{Interactional implication.} Voice interfaces should surface real-time feedback cues to reduce user hesitation.

\subsection{Misrecognition Recovery and Restart Culture}
Schegloff’s (2000) repair hierarchy labels complete restarts as the least-preferred strategy in human conversation. However, studies of AI-mediated interaction have found that, following voice-recognition failures, users often respond by issuing fresh, independently packaged commands rather than attempting repair or clarification. For example, after a system mishears a request (``Play The Bee-tulls''), users may immediately issue a more explicit command (``Play `Bohemian Rhapsody' by Queen, studio version''), adding artist, track, and rendition qualifiers to hedge against renewed error. Such behavior reflects low expectations of system memory and prioritizes efficiency over the cooperative accumulation of common ground.

\textit{Interactional implication.} Incremental confirmation (e.g., ``Did you mean the Beatles?'') can curb costly full restarts.

\subsection{Literalism and the Decline of Implicature}
Relevance Theory (Sperber \& Wilson 1986) predicts that speakers calibrate explicitness to their interlocutor’s inferential capacity. In the context of AI-mediated interactions, research has shown that users frequently abandon metaphor and idiom, opting instead for more explicit, literal language. For instance, the prompt ``Pep things up'' may result in lower system accuracy, whereas a slot-filled instruction such as ``Play an upbeat workout playlist'' achieves significantly higher success rates (Noveck \& Reboul 2008). Experimental pragmatics confirms this flattening, and Neale’s (1992) interpretation of Grice helps explain why implicature erodes when the Cooperative Principle is compromised.

\textit{Interactional implication.} Broader semantic parsing of common idioms would preserve conversational naturalness.

\subsection{Age, Learning, and Exposure Effects}
Communication Accommodation Theory posits that linguistic adjustment varies across social groups (Giles \& Ogay 2007). Existing studies indicate that age can be a salient moderator: older users are more likely to perceive bare imperatives as ``rude,'' and their satisfaction tends to decrease when politeness markers are omitted, whereas younger users often show negligible change in their perceptions or preferences. These findings underscore the need for adaptive interfaces that reflect diverse sociolinguistic comfort zones.

\textit{Interactional implication.} Optional politeness scaffolds or persona cues can accommodate tone-sensitive users.

\section{Implications and Future Directions}
The foregoing analysis shows that MFE is not a temporary workaround but a register crystallized by technological affordances and, in turn, one that feeds back into system design, pedagogy, and linguistic theory.

\subsection{Registers Under Constraint}
Halliday's Field--Tenor--Mode model (1985, 2006) presumes interlocutors who mutually negotiate meaning. MFE disrupts that assumption by introducing an algorithmic participant that neither shares experiential context nor reciprocates pragmatic effort. As a result, Field narrows toward task-oriented domains, Tenor becomes unidirectional, and Mode is delimited by parsing requirements rather than sensory channels. Accelerated enregisterment loops (Agha 2003) stabilize these conventions, inviting a broader reevaluation of how non-human interlocutors reshape the language ecology. Earlier thinking on registers under technological constraint (Halliday 1962) foreshadows this shift.

\subsection{User-Experience Design Imperatives}
MFE imposes cognitive overhead by compelling users to monitor system-legible syntax. Three design principles can alleviate that burden:
\begin{enumerate}[label=(\alph*), leftmargin=*]
    \item \textbf{Robust, naturalistic parsing}---adopt language models tolerant of paraphrase, regional variation, and mild disfluency.
    \item \textbf{Incremental, real-time feedback}---surface syntactic or semantic ambiguities as the user speaks or types, enabling in-the-moment adjustment.
    \item \parbox[t]{0.95\linewidth}{\textbf{Context-sensitive scaffolding}---present unobtrusive templates (e.g., {\small ``Set reminder for [time]''}) that guide novices without ossifying command structures.}
\end{enumerate}

\subsection{Risks for Language Learning}
Because MFE is explicit and modular in structure, it can serve as a scaffold for temporal expressions, imperative syntax, and pronunciation in second-language contexts. However, overexposure to MFE risks narrowing pragmatic competence: learners may internalize directive registers at the expense of indirectness, nuance, and rapport management. Jun (2024) and Kim \& Su (2024) show that AI tools can flatten Korean pragmatic norms, while Hossain (2021) reports similar risks in EFL contexts.

\subsection{Research Prospects and Data Needs}
To advance empirical work on MFE, scholarship will eventually need richer data resources; no dedicated, annotated corpus currently exists. Exploratory avenues include:
\begin{itemize}
    \item \textbf{Multilingual sampling}---collect illustrative MFE instances across typologically diverse languages to probe how morphology and politeness systems affect register formation.
    \item \textbf{Modality contrasts}---compare voice-only, text-only, and hybrid interactions to identify modality-specific adaptations.
    \item \textbf{Longitudinal observation}---track whether MFE conventions bleed into human--human discourse over time, signaling broader linguistic influence.
\end{itemize}

\subsection{Cross-Linguistic Reflections: Korean and English}
Typological differences mediate how MFE manifests. Korean's agglutinative structure permits omission of subjects and honorifics---yielding concise commands such as \textit{allam kkeo} (“turn off alarm”)---while retaining temporal adverbs for disambiguation. English, an analytic language, relies on word order and prepositional phrases (e.g., ``at 3 p.m.'') to achieve the same clarity. Recent systematic reviews indicate that learners of Korean in AI-mediated settings request clearer chatbot identity cues to calibrate politeness (Jeon, Lee, \& Choi 2024).

\subsection{Limitations and Tentative Next Steps}
This study is qualitative and AI-assisted; its thematic observations and reflexive drafting via NLD-P offer initial insight but do not substitute for large-scale validation. Possible follow-up efforts include:
\begin{enumerate}[leftmargin=*]
    \item A methods-focused exposition of NLD-P that clarifies prompt taxonomy and human-in-the-loop safeguards.
    \item A developer-centered, controlled evaluation comparing NLD-P with code-style and free-form prompts.
    \item Broader multimodal sampling that incorporates voice, text, and gesture interactions.
\end{enumerate}

\section{Reflexive Note on AI-Assisted Drafting}
\subsection{Disclosure Framework}
Current editorial guidelines (e.g., COPE 2023; Elsevier 2024) require authors to specify whether, and how, large language-model tools contributed to a manuscript. The following paragraphs satisfy that requirement while keeping the analytic spotlight on MFE.

\subsection{LLM Configuration and Workflow}
A private assistant---internally named ``Evalyn''---was initialized using an NLD-P system message, in which all instructions are articulated as plain-English declaratives specifying desired outcomes.

To concretize the NLD-P approach, we provide below an excerpt from the system prompt used to configure Evalyn for this research. This pseudo-coded, declarative framework defines authorial roles, editorial constraints, and operational objectives in natural language---offering a transparent alternative to developer-centric prompt formats. The full prompt specification and validation procedures will be detailed in a dedicated methods paper.

\begin{quote}
\begin{lstlisting}
[IDENT]
assistant = Evalyn
mode = scholarly_coresearcher
domain = research_assistant
type = declarative_guidance
version = 1.8
author = Hyunwoo Kim (PI)
project = R&D-2: Machine-Facing English Research
purpose = Assist a foundational research paper on 'machine-facing English'---
a linguistic phenomenon at the intersection of AI interaction and UX discourse

[RULE:AUTHORIAL_MASTERY]
writing_competence = expert_academic_level
editing_scope = macrostructure + microstyle
proofreading_focus = grammar, cohesion, syntax, factual alignment,
formatting, tone matching
...
\end{lstlisting}
{\footnotesize\textit{Excerpted for brevity. The full prompt and methodology will be presented in a subsequent paper.}}
\end{quote}

Throughout this project, we implemented a multi-tiered workflow utilizing the most advanced GPT-4-class models and Deep Research mode to ensure scholarly reliability. Primary drafting and revision were conducted with GPT-4o and GPT-4.5 (search-enabled), leveraging their advanced reasoning, rapid response, and integrated web search for high-quality academic prose and real-time citation checks.

Syntactic and stylistic refinement was accomplished using GPT-4.5, GPT-4-1 mini, and GPT-4-1, which delivered optimal performance for consistency passes, micro-edits, and stylistic harmonization.

For cross-validation and deep verification of factual claims and bibliographic leads, we employed parallel runs in Deep Research mode across GPT-4o, GPT-4.5, and GPT-4-1. This approach ensured rigorous internal fact-checking, minimized the risk of model-specific hallucinations, and provided multi-perspective analytical robustness.

Lower-tier models were not used for any substantive research drafting or validation; all critical outputs reflect the performance and capabilities of the most advanced models available at the time of writing.

\subsection{Human Oversight Protocol}
\begin{enumerate}[label=(\alph*)]
    \item Prompt issuance (author-formulated, outcome-oriented).
    \item AI generation of candidate text or bibliographic leads.
    \item Line-by-line human screening for logical coherence, stylistic consistency, and citation accuracy.
    \item Integration or rejection; most raw output served as scaffolding.
    \item Final human pass for uniform register and journal style.
\end{enumerate}
No personally identifying or proprietary data were exposed to external systems; the workflow qualifies as exempt internal research.

\subsection{Scope Boundary}
While NLD-P demonstrably shaped the writing experience, its taxonomy, validation metrics, and comparative performance lie outside the remit of this paper. A dedicated methods article will address those topics. Here, disclosure serves solely to document provenance and ensure methodological transparency without detracting from the core contribution: a conceptual account of MFE.

\section{Conclusion}
This paper has characterized Machine-Facing English as a stable register that surfaces when speakers accommodate AI processing constraints. Five linguistic features characterize MFE: redundant clarity, directive syntax, controlled vocabulary, flattened prosody, and single-intent structuring. These features were interpreted through audience design, Communication Accommodation Theory, and Gricean pragmatics; Halliday's Field--Tenor--Mode model functioned as a descriptive scaffold. Qualitative evidence from bilingual Korean- and English-language interactions illustrated how feedback loops accelerate enregisterment and reshape interactional norms.

While the primary contribution is descriptive and theoretical, two applied domains merit attention. Interface designers may draw on this analysis to lessen the cognitive overhead that nudges users toward schematic commands. Language-teaching practitioners can leverage MFE's clarity for focused drills, provided such tasks are balanced by open-ended activities that maintain pragmatic breadth.

Future work will unfold in stages: a methods article will formalize NLD-P, followed by a developer-centered, controlled evaluation of NLD-P. These incremental steps serve the broader aim of refining our understanding of how prompting practices and machine-facing registers co-evolve, ultimately clarifying how future conversational systems might accommodate---rather than necessitate---a machine-facing mode of speech.

\twocolumn
\section*{References}
{\small
\raggedright
Agha, A. (2003). The social life of cultural value. \textit{Language \& Communication}, 23(3), 231--253.\\[.6em]
Agha, A. (2004). Registers of language. In A. Duranti (Ed.), \textit{A companion to linguistic anthropology} (pp. 23--45). Blackwell.\\[.6em]
Apple. (2019). \textit{Human interface guidelines: Voice}. Apple Developer Documentation.\\[.6em]
Bell, A. (1984). Language style as audience design. \textit{Language in Society}, 13(2), 145--204.\\[.6em]
Brown, P., \& Levinson, S. C. (1987). \textit{Politeness: Some universals in language usage}. Cambridge University Press.\\[.6em]
Choi, H. (2022). Discourse strategies in Korean chatbot interactions. \textit{Korean Journal of Pragmatics}, 18(1), 45--67.\\[.6em]
Clark, H. H. (1996). \textit{Using language}. Cambridge University Press.\\[.6em]
Cohn, M., \& Zellou, G. (2021). Prosodic differences in human- and Alexa-directed speech, but similar error correction strategies. \textit{Frontiers in Communication}, 6, 675704.\\[.6em]
Cohn, M., Barreda, S., Estes, K. G., Zhou, Y., \& Zellou, G. (2024). Children and adults produce distinct technology- and human-directed speech. \textit{Scientific Reports}, 14, 15611.\\[.6em]
Giles, H., \& Ogay, T. (2007). Communication accommodation theory. In B. B. Whaley \& W. Samter (Eds.), \textit{Explaining communication: Contemporary theories and exemplars} (pp. 293--310). Lawrence Erlbaum.\\[.6em]
Google. (2020). \textit{Conversational design best practices for the Google Assistant}. Google Developers Documentation.\\[.6em]
Grandy, R. E., \& Warner, R. (2005). Paul Grice. In E. N. Zalta (Ed.), \textit{The Stanford Encyclopedia of Philosophy}.\\[.6em]
Grice, H. P. (1975). Logic and conversation. In P. Cole \& J. L. Morgan (Eds.), \textit{Syntax and semantics} (Vol. 3, pp. 41--58). Academic Press.\\[.6em]
Halliday, M. A. K. (1962). Linguistics and machine translation. \textit{STUF -- Language Typology and Universals}, 15(1--4), 145--158.\\[.6em]
Halliday, M. A. K. (1974). The place of `functional sentence perspective' in the system of linguistic description. In F. Danes (Ed.), \textit{Papers on functional sentence perspective} (pp. 43--53). Mouton.\\[.6em]
Halliday, M. A. K. (1985). \textit{An introduction to functional grammar}. Edward Arnold.\\[.6em]
Halliday, M. A. K. (2006). \textit{On language and linguistics} (Vol. 3). A\&C Black.\\[.6em]
Hossain, M. M. (2021). The application of Grice maxims in conversation: A pragmatic study. \textit{Journal of English Language Teaching and Applied Linguistics}, 3(10), 32--40.\\[.6em]
Jeon, J., Lee, S., \& Choi, S. (2024). A systematic review of research on speech-recognition chatbots for language learning: Implications for future directions in the era of large language models. \textit{Interactive Learning Environments}, 32(8), 4613--4631.\\[.6em]
Kim, S., \& Lee, J. (2020). Pragmatic markers in Korean human--computer dialogues. \textit{Journal of East Asian Linguistics}, 29(2), 201--222.\\[.6em]
Lee, S. (2019). Honorific omission in Korean voice assistant interactions. \textit{Journal of Korean Linguistics}, 35(2), 123--145.\\[.6em]
Luger, E., Sellen, A., \& Arnold, D. (2015). ``Like having a really bad PA'': The gulf between user expectation and experience of conversational agents. In \textit{Proceedings of CHI '15} (pp. 528--537).\\[.6em]
Neale, S. (1992). Paul Grice and the philosophy of language. \textit{Linguistics and Philosophy}, 15(5), 509--559.\\[.6em]
Noveck, I. A., \& Reboul, A. (2008). Experimental pragmatics: A Gricean turn in the study of language. \textit{Trends in Cognitive Sciences}, 12(11), 425--431.\\[.6em]
Pardo, J. S. (2006). On phonetic convergence during conversational interaction. \textit{The Journal of the Acoustical Society of America}, 119(4), 2382--2393.\\[.6em]
Park, J., \& Kim, Y. (2021). Pragmatic simplification in chatbot interactions among elderly Korean users. \textit{International Journal of Human--Computer Studies}, 150, 102640.\\[.6em]
Porcheron, M., Fischer, J. E., Reeves, S., \& Sharples, S. (2018). Voice interfaces in everyday life: Examining human--AI interaction in the wild. In \textit{Proceedings of CHI '18} (Paper 640).\\[.6em]
Reitter, D., Moore, J. D., \& Keller, F. (2006). Predicting success in dialogue. In \textit{Proceedings of ACL '06} (pp. 311--318).\\[.6em]
Sacks, H., Schegloff, E. A., \& Jefferson, G. (1974). A simplest systematics for the organization of turn taking for conversation. \textit{Language}, 50(4), 696--735.\\[.6em]
Schegloff, E. A. (2000). Overlapping talk and the organization of turn taking for conversation. \textit{Language in Society}, 29(1), 1--63.\\[.6em]
Schwab, S., \& Zellou, G. (2020). Child-directed clear speech: Acoustic phonetic changes in caregivers' speech to children with typical and atypical language development. \textit{Journal of Speech, Language, and Hearing Research}, 63(9), 2974--2987.\\[.6em]
Searle, J. R. (1969). \textit{Speech acts: An essay in the philosophy of language}. Cambridge University Press.\\[.6em]
Song, J. Y., Pycha, A., \& Culleton, T. (2022). Interactions between voice-activated AI assistants and human speakers and their implications for second language acquisition. \textit{Frontiers in Communication}, 7, 995475.\\[.6em]
Sperber, D., \& Wilson, D. (1986). \textit{Relevance: Communication and cognition}. Harvard University Press.\\[.6em]
Van Engen, K. J., \& Bradlow, A. R. (2007). Native language effects in second language listening: The case of clear speech. \textit{The Journal of the Acoustical Society of America}, 121(1), 553--562.\\[.6em]
Ward, N., \& Tsukahara, W. (2000). Prosodic features that cue back channel responses in English and Japanese. \textit{Journal of Pragmatics}, 32(8), 1177--1207.\\[.6em]
}
\end{document}